\documentclass{article}

\usepackage[square,sort,comma,numbers]{natbib}
\usepackage[final]{style}

\usepackage[utf8]{inputenc} 
\usepackage[T1]{fontenc}    
\usepackage{url}            
\usepackage{booktabs}       
\usepackage{amsfonts}       
\usepackage{nicefrac}       
\usepackage{microtype}      
\usepackage{xcolor}         
\usepackage{graphicx}
\usepackage{amsmath}
\usepackage{amssymb}
\usepackage{wrapfig,lipsum,booktabs} 
\usepackage{multirow}
\usepackage{makecell}
\usepackage{nicefrac} 
\usepackage{xcolor}         
\usepackage{algorithm}
\usepackage{algorithmic}
\usepackage{url}
\usepackage{makecell}
\usepackage[pagebackref,breaklinks,colorlinks]{hyperref}

\usepackage{algorithm}
\usepackage{algorithmic}
\newtheorem{theorem}{Theorem}

\newtheorem{lemma}{Lemma}

\title{Generalization Capability for Imitation Learning}

\author{%
  Yixiao Wang \\
  University of California, Berkeley\\
  Berkeley, CA 94704 \\
\texttt{yixiao\_wang@berkeley.edu} \\
}

\begin{document}
\maketitle

\begin{abstract}
Imitation learning holds the promise of equipping robots with versatile skills by learning from expert demonstrations. However, policies trained on finite datasets often struggle to generalize beyond the training distribution. In this work, we present a unified perspective on the generalization capability of imitation learning, grounded in both \emph{information theorey}  and \emph{data distribution property}. We first show that the generalization gap can be upper bounded by (i) the conditional information bottleneck on intermediate representations and (ii) the mutual information between the model parameters and the training dataset. This characterization provides theoretical guidance for designing effective training strategies in imitation learning, particularly in determining whether to freeze, fine-tune, or train large pretrained encoders (e.g., vision-language models or vision foundation models) from scratch to achieve better generalization. Furthermore, we demonstrate that high conditional entropy from input to output induces a flatter likelihood landscape, thereby reducing the upper bound on the generalization gap. In addition, it shortens the stochastic gradient descent (SGD) escape time from sharp local minima, which may increase the likelihood of reaching global optima under fixed optimization budgets. These insights explain why imitation learning often exhibits limited generalization and underscore the importance of not only scaling the diversity of input data but also enriching the variability of output labels conditioned on the same input.
\end{abstract}

\section{Introduction}
Learning robot policies directly from demonstrations—\emph{imitation learning}—has emerged as a powerful paradigm for scaling robotic capabilities. The imitation learning problem can be formulated as learning a robot’s future action $Y$ based on the available input information $X$ (e.g., historical images, task descriptions in natural language, the robot’s current state, etc.). Given a dataset $s = \{(x_i, y_i)\}_{i=1}^n$, where $x_i \in \mathcal{X}$ and $y_i \in \mathcal{Y}$, drawn from a joint distribution $(X, Y) \sim \mathcal{P}$, the goal is to learn a neural network $f^s(\cdot)$ that approximates the mapping $y = f^s(x)$.

While large networks are capable of fitting large collections of demonstrations, the true deployment distribution $\mathcal{P}(X, Y)$ is typically far broader than the support of the finite dataset $s$, especially for robot tasks. This mismatch leads to a potentially significant and safety-critical \emph{generalization gap}. Recent research \cite{lin2024data, khazatsky2024droid, black2024pi0visionlanguageactionflowmodel, bjorck2025gr00t} has demonstrated that scaling robot datasets, mixing them with internet-scale data, incorporating cross-embodiment demonstrations, and leveraging pretrained foundation models can improve the generalization capability of learned policies. However, the choice of training strategies remains unclear and is largely determined through empirical evaluation. For example, \cite{huang2025otter} shows that fine-tuning a CLIP \cite{radford2021learning} encoder on robot datasets can reduce generalization to unseen objects, and \cite{bjorck2025gr00t} also uses a frozen vision-language model (VLM) encoder. In contrast, \cite{black2024pi0visionlanguageactionflowmodel} fine-tunes the entire model on the robot dataset. The underlying mechanisms that influence generalization in imitation learning remain poorly understood, and it is crucial to provide mathematical insight and principled guidance for effective model design under varying conditions.

In this paper, we investigate the generalization capability of imitation learning from both an \emph{information-theoretic} and a \emph{data distribution} perspective, aiming to reveal the key factors that govern generalization performance.

\subsection{Main Findings}
We summarize our theoretical findings as follows:
\begin{enumerate}
    \item Compressing the intermediate representation of input $X$, while retaining sufficient information to predict the output $Y$, reduces the upper bound on the generalization gap.
    \item Using encoders that are less dependent on the specific training dataset, yet still capable of achieving low training loss, leads to a smaller generalization gap.
    \item A larger conditional entropy $H(Y | X)$ provably results in a flatter likelihood landscape, which in turn reduces the dependence between the encoder and the dataset, tightening the generalization bound.
    \item A larger conditional entropy $H(Y | X)$ also accelerates the escape time of stochastic gradient descent (SGD) from sharp local minima, potentially improving optimization outcomes.
\end{enumerate}

\subsection{Actionable Guidelines}
We distill the above theoretical results into practical guidelines for improving generalization in imitation learning:
\begin{enumerate}
    \item Encourage compression in representations not only for high-dimensional modalities such as images but also for low-dimensional inputs such as proprioceptive states, especially when the dataset size is limited.
    \item Freezing or lightly finetuning large pretrained encoders can reduce the generalization gap, provided that the training loss remains comparable to that achieved by fine-tuning.
    \item It is important to increase not only the diversity of input states $X$ (i.e., task or environment diversity) but also the diversity of actions $Y$ conditioned on the same input, to promote better generalization.
\end{enumerate}

\section{Latent and Model Compressions Reduce Generalization Gap}
\subsection{Definition of Generalization Gap}
\begin{figure}[H]
\centering
\begin{minipage}{0.4\linewidth}
  \centering
  \includegraphics[width=\linewidth]{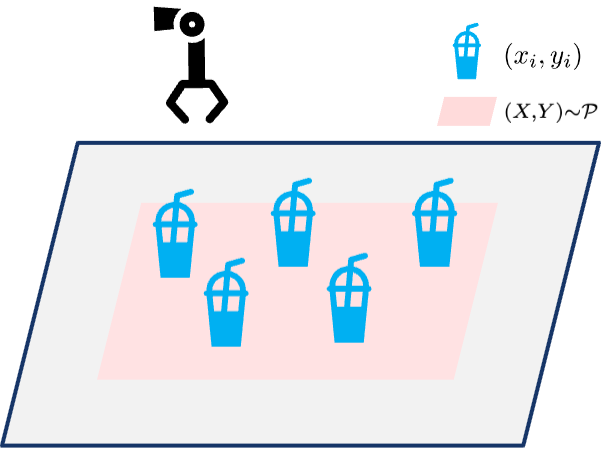}
  \caption{Illustration on generalization gap in imitation learning. We would like to utilize finite number of data $\{(x_i,y_i)\}_{i=1}^n$ and generalize to the whole distribution $(X,Y)\sim \mathcal{P}$.}
  \label{fig: generalization gap}
\end{minipage}
\hfill
\begin{minipage}{0.5\linewidth}
  \centering
  \includegraphics[width=\linewidth]{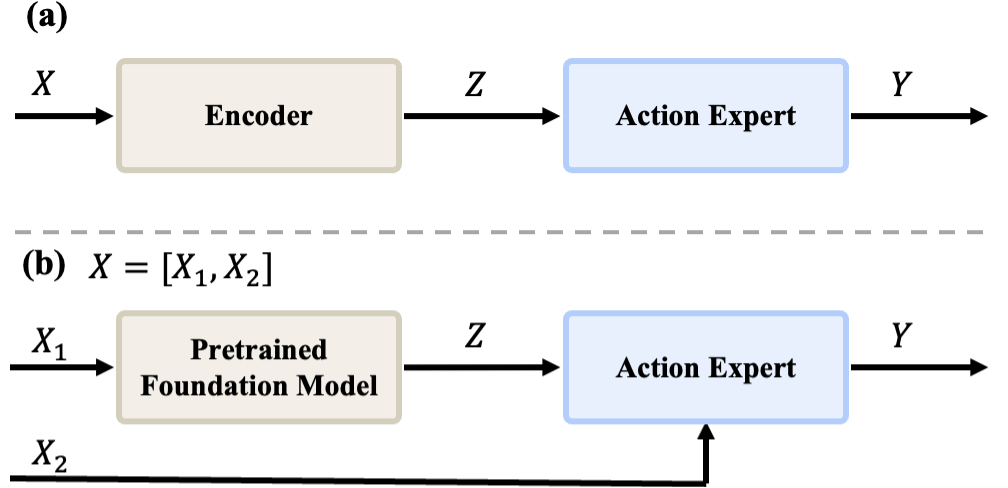}
  \caption{Common framework for imitation learning. \textbf{(a)} shows encoder all the information of $X$ together; \textbf{(b)} shows encoder large portion of $X$ such as image and lanuage through a large model, then together with intermediate representation $X$ with the rest of $X$, to generate the robot action $Y$.}
  \label{fig: rfm structure}
\end{minipage}
\end{figure}

Since the dataset may not cover the whole space of $\mathcal{P}$ as shown in Figure \ref{fig: generalization gap}, there exist the generalization gap $\Delta(s)$:
\begin{equation}
    \label{eq: generalization gap}
    \begin{aligned}
        \Delta(s):=&\mathbb{E}_{(X,Y)\sim \mathcal{P}}[\ell(f^s(X),Y)]-\frac{1}{n}\sum_{i=1}^n \ell(f^s(x_i),y_i)\\
    \end{aligned}
\end{equation}
where $\ell(\cdot)$ denotes the bounded per-sample loss (e.g., behavior cloning loss in the imitation learning domain). The generalization gap $\Delta(s)$ can be significant when the dataset does not cover the whole space of $\mathcal{P}$—a common scenario in the robot learning domain. Therefore, it is crucial to analyze the generalization gap: how well the learned policy $f^s(\cdot)$ generalizes to a wide range of tasks given only limited data.


\subsection{Problem Statement}
Since $X$ in imitation learning contains multimodal information—such as images and language, which are high-dimensional—and the robot’s proprioceptive state, which is low-dimensional, we denote $X = [X_1, X_2]$, where $X_2$ represents the robot’s proprioceptive state and $X_1$ represents the rest of $X$. Current learning-based robotic models typically follow two main frameworks, as illustrated in Figure~\ref{fig: rfm structure}:
\begin{itemize}
    \item[(a)] Process $X_1$ and $X_2$ in a nearly equal manner (e.g., through simple projections or small encoders~\cite{team2024octo}).
    \item[(b)] Encode high-dimensional modalities such as images and language using large pretrained foundation models—such as vision or vision-language models—to extract representations $Z$, which are then combined with the robot’s proprioceptive state $X_2$ to generate actions.
\end{itemize}

In both frameworks, it is crucial to understand how internal representations affect robot performance and the generalization gap $\Delta(s)$. In particular, the second framework warrants deeper investigation, as it offers multiple design choices—such as fine-tuning, freezing, or training from scratch—each suited to different tasks and performance objectives.

To formally investigate this, we define a general form of the robot policy $f^s(\cdot)$, modeled as a multilayer network (e.g., a ResNet+Transformer or a pure Transformer). For clarity, we do not explicitly separate $X_1$ and $X_2$ in the analysis and instead treat $X$ as a whole; however, the formulation naturally extends to the separated case. We analyze its internal representations, specifically the intermediate random variables at various layers. We define the intermediate representation at layer $l$ as $Z_l^s$:
\begin{equation}
    \label{eq: definition intermediate representation}
    Z_l^s = \phi_l^s(X),
\end{equation}
where $\phi_l^s(\cdot)$ denotes the subnetwork from the input to layer $l$, $l=\{1,2,...,L+1\}$. Note that $\phi_{L+1}^s(\cdot)=f^s(\cdot)$.

\subsection{Upper Bound of Generalization Gap}
We first consider the case where $\phi_l^s(\cdot)$ is independent of the training dataset $s$. This scenario commonly arises when a frozen pretrained foundation model, such as a vision foundation model (VFM) or vision-language model (VLM), is used to process the input $X$ (e.g., image or language), and an action head is subsequently applied to decode the representation $Z_l^s$ into the output action $Y$. Many works \cite{bjorck2025gr00t, majumdar2023we, huang2025otter} follow this framework. 

Theorem~\ref{theorem: encoder freeze} shows that in such case, the generalization gap is upper bounded, and this bound can be tightened when the conditional mutual information $I(X; Z_l^s | Y)$ is small. Therefore, reducing $I(X; Z_l^s | Y)$ can help improve generalization. This insight aligns with the principle of the information bottleneck, which has been widely studied in prior work \cite{lee2023conditional, jiang2024correlation, lu2020dynamics, goyal2019infobot}. These works aim to find a minimal yet sufficient representation $Z_l^s$ of the input $X$ for predicting $Y$. Experimental results support that generalization capability can be improved by applying the information bottleneck principle—specifically, by minimizing $I(X; Z_l^s | Y)$ or its equivalent decomposition $-I(Z_l^s; Y) + I(Z_l^s; X)$.

\begin{theorem}
\label{theorem: encoder freeze}
\cite{kawaguchi2023does} If $\phi_l^s(\cdot)$ is independent with the training dataset $s$, then for any $\delta>0$, with the probability at least $1-\delta$, the following holds:
\begin{equation}
    \Delta(s)\leq G_3^l \sqrt{\frac{1}{n}(I(X;Z_l^s|Y)\ln(2)+\mathcal{G}_2^l)}+\frac{G_1^l(0)}{\sqrt{n}}
\end{equation}
where $I(\cdot)$ is the mutual information,$G_1^l(0)=\Tilde{\mathcal{O}}(1)$, $G_3^l(0)=\Tilde{\mathcal{O}}(1)$, $\mathcal{G}_2^l=\Tilde{\mathcal{O}}(1)$ when $n\rightarrow \infty$, $l\in\{1,2,...,L\}$.
\end{theorem}

Though we have seen using information bottleneck can reduce the generalization gap, the assumption that $\phi_l^s(\cdot)$ is independent with the dataset $s$ is too strong. When we train our policy from scratch, or finetune the VLM and VFM, the $\phi_l^s(\cdot)$ is dependent on the dataset $s$. Theorem \ref{theorem: encoder unfreeze} shows more general case when there is no assumption that $\phi_l^s(\cdot)$ is independent with $s$. It shows that when the encoder $\phi_l^s(\cdot)$ can less dependent on the training dataset $s$, the generalization gap can be reduced. Otherwise, the generalization capability will be worse, which has been commonly seen the case that finetuning the large VLM and VFM encoder and training the encoder from scratch in the robot foundation model \cite{huang2025otter}.

\begin{theorem}
\label{theorem: encoder unfreeze}
\cite{kawaguchi2023does} For any $\delta>0$, with the probability at least $1-\delta$, the following holds:
\begin{equation}
    \begin{cases}
        \Delta(s)\leq \min_{l \in \{1,2,...,L+1\}} Q_l,\\
        Q_l = G_3^l \sqrt{\frac{1}{n}((I(X;Z_l^s|Y)+I(\phi_l^S;S))\ln(2)+\hat{\mathcal{G}}_2^l)}+\frac{G_1^l(\zeta)}{\sqrt{n}}, &l \leq L \\
        Q_l = \mathcal{R}(f^s)\sqrt{\frac{1}{2n}(I(\phi_l^S;S)\ln(2)+\check{\mathcal{G}}_2^l)}, &l == L+1\\
    \end{cases}
\end{equation}
where $S$ is the random variable representing the distribution of training dataset $s$, $G_1^l(\zeta)=\Tilde{\mathcal{O}}(\sqrt{I(\phi_l^S;S)+1})$, $\hat{\mathcal{G}}_2^l=\Tilde{\mathcal{O}}(1)$, $\check{\mathcal{G}}_2^l=\Tilde{\mathcal{O}}(1)$, $G_3^l=\Tilde{\mathcal{O}}(1)$ when $n\rightarrow \infty$.
\end{theorem}

\subsection{Conclusion}
In order to reduce the generalization gap, we can:
\begin{itemize}
    \item Enforce the compression of the latent representation $Z_l^s$ from states $X$ in the robot foundation model, while retaining sufficient information to predict the action $Y$—that is, reduce $I(X;Z_l^s|Y)$.
    \item Enforce the disentanglement between the encoder $\phi_l^s(\cdot)$ and the training dataset $s$—that is, reduce $I(\phi_l^s; S)$.
\end{itemize}

However, a smaller generalization gap does not necessarily imply good robot performance on both in-distribution and out-of-distribution tasks. Simply applying the information bottleneck principle or using a pretrained encoder does not guarantee strong performance. 

What we ultimately care about is the expected performance $\mathbb{E}_{(X,Y)\sim \mathcal{P}}[\ell(f^s(X), Y)]$ across the entire data distribution:
\begin{equation}
    \mathbb{E}_{(X,Y)\sim \mathcal{P}}[\ell(f^s(X), Y)] = \Delta(s) + \frac{1}{n} \sum_{i=1}^n \ell(f^s(x_i), y_i)
\end{equation}

Therefore, we must keep the training loss $\frac{1}{n} \sum_{i=1}^n \ell(f^s(x_i), y_i)$ very low, while simultaneously reducing $I(X;Z_l^s|Y)$ and $I(\phi_l^s; S)$. In most cases, using a pretrained VFM or VLM cannot achieve very low training loss, and applying the information bottleneck often leads to a trade-off with higher training loss.

\section{Data Distribution Properties Influence Generalization Capability}
In many generative tasks, such as text-to-image generation \cite{rombach2022high}, models often exhibit strong generalization capabilities. For instance, a model may generate an image depicting “a man on a horse eating a hamburger,” even if such an image does not appear in the training dataset. In contrast, such generalization is rarely observed in imitation learning.

A key observation is that imitation learning typically corresponds to a \emph{many-to-one} mapping, as different states often result in the same action. This stands in contrast to text-to-image generation, which exemplifies a \emph{one-to-many} mapping. Notably, a \emph{one-to-many} mapping is associated with high conditional entropy $H(Y | X)$, whereas a \emph{many-to-one} mapping corresponds to low conditional entropy. This suggests a potential link between the conditional entropy $H(Y | X)$ and the generalization gap.

In the following section, we investigate how the conditional entropy $H(Y | X)$ influences the generalization gap.

\subsection{Effect on Upper Bound of Generalization Gap}
As shown in Theorem~\ref{theorem: encoder freeze}, the generalization gap is related to $I(X; Z_l^s | Y)$ and $I(\phi_l^S; S)$. Since the conditional entropy $H(Y | X)$ is not directly related to $I(X; Z_l^s | Y)$—as $I(X; Z_l^s | Y)$ can always be compressed—we further investigate the connection between $H(Y | X)$ and $I(\phi_l^S; S)$.

Theorem~\ref{theorem: phi upper bound} shows that the upper bound of $I(\phi_l^S; S)$ is positively correlated with the trace of the Hessian matrix, $\operatorname{tr}(\mathcal{H})$. The matrix $\mathcal{H}$ can be approximated by the Fisher information matrix $F(\theta)$ around a local minimum, where $F(\theta)$ is defined as:

\begin{equation}
    F(\theta) = \mathbb{E}_{(x,y) \sim p(x) p_\theta(y|x)}[\nabla_\theta \log p_\theta(y|x)\nabla_\theta \log p_\theta(y|x)^T]
\end{equation}
where $\nabla_\theta \log p_\theta(x,y)=\nabla_\theta \log p_\theta(y|x)p(x)=\nabla_\theta \log p_\theta(y|x)$

\begin{theorem}
\label{theorem: phi upper bound} \cite{achille2018emergence}
Denote $\theta$ is the network weight of encoder $\phi_l^S$. Assume after training, $\theta$ reaches a local minimum $\hat{\theta}$, and denote $\mathcal{H}$ be the Hessian at that point. Then, for the optimal choice of the posterior $\theta | S = \epsilon \odot \hat{\theta}$ centered at $\hat{\theta}$ that optimizes the IB Lagrangian, we have
\begin{equation}
I(\theta; \mathcal{D}) \leq \frac{1}{2} K \left[\log \|\hat{\theta}\|_2^2 + \log \operatorname{tr} (\mathcal{H}) - K \log(K^2 \beta / 2)\right]
\end{equation}
where $K = \dim(\theta)$, $\operatorname{tr}(\cdot)$ is the trace of a matrix, $\beta$ is a Lagrange multiplier implicitly induced by stochastic gradient descent under certain conditions \cite{achille2018emergence, chaudhari2018stochastic}.
\end{theorem}

In order to reduce the Fisher information $F(\theta)$, Theorem~\ref{theorem: trace fisher upper bound} shows that one can decrease the KL divergence between the uniform distribution and the conditional distribution $p(y | x)$, denoted as $D_x$, thereby lowering $F(\theta)$. Note that for a fixed finite-volume support, the uniform distribution achieves the highest entropy. This implies that a higher conditional entropy $H(Y | X)$ leads to a smaller $D_x$, which in turn results in a lower trace of the Fisher information matrix, $\operatorname{tr}(F(\theta))$. Consequently, this corresponds to a flatter likelihood surface and a lower upper bound on the mutual information $I(\theta; S)$.

\begin{lemma}
\label{lemma: probability bound}
Assume the conditional density $p(y | x)$ has support $\mathcal{Y}_x \subset \mathbb{R}^{m}$ with finite volume $V_x := \operatorname{vol}(\mathcal{Y}_x)$. Define its \emph{entropy gap} as
\begin{equation}
    D_x := \log V_x - H(Y | X = x)
         = \mathrm{D}_{\mathrm{KL}}\!\bigl(p(\cdot | x) \,\Vert\, u_x\bigr),
    \qquad
    u_x(y) \equiv 1/V_x,
\end{equation}
where $u_x$ is the uniform distribution over $\mathcal{Y}_x$.
By Pinsker’s inequality,
\begin{equation}
\sup_{y\in\mathcal Y_x}\!
\bigl|p(y| x)-u_x(y)\bigr|
\;\le\;
\sqrt{\tfrac{D_x}{2}}.
\end{equation}
Consequently, for any $y_1,y_2\in\mathcal Y_x$,
\begin{equation}
\bigl|p(y_1| x)-p(y_2| x)\bigr|
\;\le\;
\sqrt{2D_x}.
\end{equation}    
\end{lemma}

\begin{theorem}
\label{theorem: trace fisher upper bound}
Assume the learned model $p_\theta(y | x)$ is well-trained, i.e., $p_\theta(y | x) \approx p(y | x)$, and satisfies the following conditions:
\begin{enumerate}
    \item \textbf{Likelihood Range (Lemma \ref{lemma: probability bound}).}\quad
          $\displaystyle
          \bigl|p(y_1| x)-p(y_2| x)\bigr|
            \;\le\;
            \sqrt{2D_x} \quad \text{for any } y_1, y_2 \in \mathcal{Y}_x$
    \item \textbf{Minimal likelihood.}\quad
          $\displaystyle
          p_\theta(y | x) \ge \epsilon_x
          \quad \text{for all } y \in \mathcal{Y}_x$
          for some $\epsilon_x > 0$;
\end{enumerate}

Consider Lemma \ref{lemma: probability bound}, then, for a sufficiently small and fixed perturbation $\delta\theta$, the score vector can be approximately bounded as
\begin{equation}
    |\nabla_\theta \log p_\theta(y | x)|
    \;\lesssim\;
    C_x := \frac{1}{\delta \theta} \log\!\left(1 + \frac{\sqrt{2 D_x}}{\epsilon_x}\right)
    \quad \text{for all } y \in \mathcal{Y}_x.
\end{equation}

Consequently, the Fisher information matrix satisfies the following trace bounds:
\begin{equation}
   \operatorname{tr} F(\theta) \le \max_x [C_x^TC_x].
\end{equation} 
\end{theorem}

\subsection{Effect on Training Dynamics}

We further investigate how the conditional entropy $H(Y | X)$ influences the training dynamics with stochastic gradient descent (SGD) optimizer. Theorem~\ref{theorem:sgd-escape} shows that the mean escape time from a local minimum is exponentially related to the loss difference $\Delta \mathcal{L} = \mathcal{L}(b) - \mathcal{L}(a)$. From Lemma~\ref{lemma: probability bound}, we obtain that $\Delta \mathcal{L}$ is upper bounded by $D_x$. 

As discussed earlier, a higher conditional entropy $H(Y | X)$ implies a smaller $D_x$, which in turn leads to a smaller $\Delta \mathcal{L}$ and thus a shorter escape time. A lower escape time increases the likelihood that the optimizer escapes poor local minima and reaches a global optimum within a fixed number of optimization steps. Therefore, a higher $H(Y | X)$ not only facilitates achieving lower training loss but also increases the potential for the optimizer to converge to a generalizable solution that better captures the true data distribution $\mathcal{P}$.

\begin{theorem}\cite{xie2020diffusion}
\label{theorem:sgd-escape}
The loss function $\mathcal{L}(\theta)$ is of class $C^2$ and $n$-dimensional. Only one most possible path exists between Valley~$a$ and the outside of Valley~$a$. If Assumption \textcolor{red}{1}, \textcolor{red}{2}, and \textcolor{red}{3} in \cite{xie2020diffusion} hold, and the dynamics is governed by SGD, then the mean escape time from Valley~$a$ to the outside of Valley~$a$ is
\[
\tau = 2\pi \frac{1}{|\mathcal{H}_{be}|} \exp\left[ \frac{2B\Delta \mathcal{L}}{\eta} \left( \frac{p}{\mathcal{H}_{ae}} + \frac{(1 - p)}{|\mathcal{H}_{be}|} \right) \right],
\]
where $p \in (0,1)$ is a path-dependent parameter, and $\mathcal{H}_{ae}$ and $\mathcal{H}_{be}$ are, respectively, the eigenvalues of the Hessians at the minimum $a$ and the saddle point $b$ corresponding to the escape direction $e$, $\Delta \mathcal{L}=\mathcal{L}(b)-\mathcal{L}(a)$, $B$ is the batch size.
\end{theorem}

\subsection{Conclusion}
In this section, we have shown that, in general, a high conditional entropy $H(Y | X)$ leads to a lower upper bound on the mutual information $I(\theta; S)$, which in turn lowers the upper bound of the generalization gap $\Delta(s)$. In addition, a high $H(Y | X)$ results in shorter escape times from local minima, thereby facilitating convergence to global minima. Finding a global minimum not only reduces the training loss but also increases the likelihood of achieving lower loss across the true data distribution $\mathcal{P}$. This may explain why strong generalization is often observed in text-to-image generation tasks, yet remains elusive in the domain of robot policy learning.

In robot foundation models, the objective is to train a policy that can both accurately and generalizably perform diverse tasks, often using high-quality expert demonstrations. In many existing datasets, given an observation and task description $X$, the corresponding robot action $Y$ is nearly deterministic. This is because most robot datasets are collected using either off-the-shelf algorithms or human teleoperation. When using off-the-shelf algorithms—especially optimization-based motion planners—the action $Y$ is largely determined by the current state $X$. Even with sampling-based planners, once the environment is fixed, the feasible action space $Y$ becomes quite constrained. Similarly, when using human teleoperation, the demonstrated actions are often biased by habitual patterns, leading again to a nearly deterministic mapping from $X$ to $Y$.

Therefore, in order to improve generalization capability in imitation learning, it becomes necessary to explicitly constrain $I(\theta; S)$ when the dataset does not exhibit high $H(Y | X)$. In contrast to other domains—where the high entropy of the dataset naturally imposes a lower upper bound on $I(\theta; S)$—robot policy learning requires either the enforcement of regularization techniques to reduce $I(\theta; S)$ or a careful reconsideration of dataset design. Specifically, one may need to curate or collect datasets that encourage higher $H(Y | X)$.

One might argue that using a video prediction model followed by an inverse dynamics module could enhance generalization. However, it is important to note that the informational content of the next video frame is essentially equivalent to that of the next action. Therefore, in this context, there is no fundamental distinction between next-frame prediction and action prediction. Furthermore, if one employs a pretrained video generation model and subsequently fine-tunes it on robotic video data, this scenario falls outside the scope of our current discussion. Such pretrained models inherently incorporate prior knowledge from external datasets, which typically exhibit high conditional entropy $H(Y | X)$. As a result, the mutual information $I(\theta; S)$ is naturally constrained to be lower—unless the model overfits to the fine-tuning dataset $s$.

\section{Conclusion} \label{sec:conclusion}
In this paper, we study on why imitation policies struggle to generalize and how they can be improved. We first analyze the upper bound of the generalization gap, highlighting two critical factors: \emph{representation compression} and \emph{encoder–dataset dependence}, which are quantitatively captured by $I(X; Z_l^{s} | Y)$ and $I(\phi_l^{S}; S)$, respectively. Next, we investigate the role of action stochasticity. Specifically, we show that high conditional entropy $H(Y | X)$ reduces the Fisher information trace, thereby lowering the upper bound of the generalization gap. Moreover, it shortens the stochastic gradient descent (SGD) escape time from sharp local minima, increasing the likelihood of reaching globally optimal solutions. In contrast, deterministic many-to-one mappings—typical of robotic datasets—leave the generalization gap largely unconstrained. To improve generalization in practice, it is essential to maintain low training loss while simultaneously: (i) enforcing compression on both representations and network weights, and (ii) redesigning data collection protocols to inject controlled stochasticity into the action space, beyond merely increasing input state diversity.

Our analysis suggests that progress in generalist robot policies cannot rely on data scaling alone; it also demands \emph{information-aware architectures, optimization strategies, and dataset design}. Future work will empirically validate these findings and extend the theory to the post-training stage, including the incorporation of reward signals from either the environment or human feedback.

\bibliographystyle{plain}
\bibliography{main}


\end{document}